# Generalized sequential tree-reweighted message passing

**Vladimir Kolmogorov**
Institute of Science and Technology Austria
vnk@ist.ac.at

**Thomas Schoenemann**
Heinrich-Heine-Universität Düsseldorf
tosch@phil.uni-duesseldorf.de

## Abstract

This paper addresses the problem of approximate MAP-MRF inference in general graphical models. Following [36], we consider a family of linear programming relaxations of the problem where each relaxation is specified by a set of nested pairs of factors for which the marginalization constraint needs to be enforced. We develop a generalization of the TRW-S algorithm [9] for this problem, where we use a decomposition into *junction chains, monotonic w.r.t. some ordering on the nodes*. This generalizes the *monotonic chains* in [9] in a natural way. We also show how to deal with nested factors in an efficient way. Experiments show an improvement over min-sum diffusion, MPLP and subgradient ascent algorithms on a number of computer vision and natural language processing problems.

## 1 Introduction

This paper is devoted to the problem of minimizing a function of discrete variables represented as a sum of *factors*, where a factor is a term depending on a certain subset of variables. The problem is also known as MAP-MRF inference in a graphical model. Due to the generality of the definition, it has applications in many areas. Probably, the most well-studied case is when each factor depends on at most two variables (*pairwise MRFs*). Many inference algorithms have been proposed. One prominent approach is to try to solve a natural linear programming (LP) relaxation of the problem, sometimes called *Schlesinger LP* [35]. A lot of research went into developing efficient solvers for this special LP; some example are [34, 9, 35, 12, 5, 8, 20, 27, 2, 30, 15, 17, 22].

A similar LP can also be formulated for higher-order MRFs. In fact, this can be done in many ways. We follow the formalism of [36] who describes a family of LP relaxations specified by a set of pairs of nested factors for which the marginalization constraint needs to be enforced. This approach can also be used for pairwise MRFs: we can obtain a hierarchy of progressively tighter relaxations by (i) grouping some pairwise factors into larger factors (or introducing higher-order factors with zero cost functions), and (ii) formulating an LP for the resulting higher-order MRF. This hierarchy covers the *Sherali-Adams hierarchy* but gives a finer control over the relaxation (see [26]).

**Contributions** We present a new algorithm for solving the relaxation discussed above. It builds on the *sequential tree-reweighted message passing* (TRW-S) algorithm of [9] (which in turn builds on [34]). TRW-S showed a good performance for pairwise MRFs [29, 30, 22], so generalizing it to higher-order MRFs is a natural direction. While developing such a generalization, we had to overcome some technical difficulties such as finding the right definition for *monotonic junction chains* and deciding how to deal with nested factors.

**Related work** A general framework for obtaining convergent algorithms called *tree-consistency bound optimization* (TBCO) was proposed in [16]. It covers many existing techniques (such as *MSD* and *MPLP*), as well as ours. However, the authors of [16] did not propose any specific choices for the case of higher-order factors, restricting their experiments to 4-connected grids. The efficiency of computing min-marginals was also not considered. In contrast, the focus of our paper is on investigating which choices lead to more efficient techniques in practice. Note, monotonicity for the higher-order case was not mentioned in [16].

Another related technique is the *min-sum diffusion* (MSD) algorithm [36]. It can be shown they have similar theoretical properties; they both monotonically increase a lower bound on the function, and are characterized by similar stopping criteria. Note, they are not guaranteed to solve the LP exactly - they may get stuck in a suboptimal point [9, 35]. Other techniques with such properties (formulated for restricted cases) include MPLP [25, 27] and the method in [14]; they address the problem of tightening Schlesinger LP for pairwise MRFs. [6] considered the case of *factor graphs*, or relaxations with *singleton separators*.

A lot of research went into developing algorithms that are guaranteed to converge to an optimal solution of the LP. Examples include subgradient ascent techniques ([10, 11]), proximal projections ([20]), Nesterov schemes ([8, 21]), an augmented Lagrangian method [15, 17], and the technique in [22] described as the "smoothed version of TRW-S". According to [22], the latter outperforms many other techniques on the stereo problem.

Our results in section 5 indicate that TRW-S generally outperforms other popular techniques that we tested, namely MSD, MPLP and a subgradient ascent.

## 2 Background and notation

We will closely follow the notation of [36]. Let $V$ be the set of nodes. For each node $v \in V$ let $\mathcal{X}_v$ be the finite set of possible labels for $v$, and $\mathcal{X} = \otimes_{v \in V} \mathcal{X}_v$ be the set of labelings of $V$. Our goal will be to minimize the function

$$f(\boldsymbol{x} \mid \bar{\theta}) = \sum_{A \in \mathcal{F}} \bar{\theta}_A(\boldsymbol{x}_A), \quad \boldsymbol{x} \in \mathcal{X} \tag{1}$$

where $\mathcal{F} \subset 2^V$ is a set of non-empty subsets of $V$ (also called *factors*), $\boldsymbol{x}_A$ is the restriction of $\boldsymbol{x}$ to $A \subseteq V$, and $\bar{\theta}$ is a vector with components $(\bar{\theta}_A(\boldsymbol{x}_A) \mid A \in \mathcal{F}, \boldsymbol{x}_A \in \otimes_{v \in A} \mathcal{X}_v)$.

Let $J$ be a fixed set of pairs of the form $(A, B)$ where $A, B \in \mathcal{F}$ and $B \subset A$. Note that $(\mathcal{F}, J)$ is a directed acyclic graph. We will be interested in solving the following relaxation of the problem:

$$\min_{\boldsymbol{\mu} \in \mathcal{L}(J)} \sum_{A \in \mathcal{F}} \sum_{\boldsymbol{x}_A} \bar{\theta}(\boldsymbol{x}_A) \mu_A(\boldsymbol{x}_A) \tag{2}$$

where $\mathcal{L}(J)$ is the $J$-based *local polytope* of $(V, \mathcal{F})$:

$$\mathcal{L}(J) = \left\{ \boldsymbol{\mu} \geq 0 \;\middle|\; \begin{array}{l} \sum_{\boldsymbol{x}_A} \mu_A(\boldsymbol{x}_A) = 1 \quad \forall A \in \mathcal{F}, \boldsymbol{x}_A \\ \sum_{\boldsymbol{x}_{A-B}} \mu_A(\boldsymbol{x}_A) = \mu_B(\boldsymbol{x}_B) \\ \qquad \forall (A, B) \in J, \boldsymbol{x}_B \end{array} \right\} \tag{3}$$

Here and below we use the following implicit restriction convention: for $B \subseteq A$, whenever symbols $\boldsymbol{x}_A$ and $\boldsymbol{x}_B$ appear in a single expression they do not denote independent joint states but $\boldsymbol{x}_B$ denotes the restriction of $\boldsymbol{x}_A$ to nodes in $B$.

As an example, one could define $J = \{(A, \{v\}) \mid A \in \mathcal{F}, v \in A\}$; graph $(\mathcal{F}, J)$ is then known as a *factor graph*. It can be shown that the resulting relaxation is tight if each term $\bar{\theta}_A$ is a submodular function [36], but for non-submodular functions we may need to add extra edges to $J$ to tighten the relaxation. Note, in general conditions $A, B \in \mathcal{F}, B \subseteq A$ don't imply that $(A, B) \in J$. Requiring the latter would be unreasonable; if, for example, $|A|, |B| \gg 1$ then adding edge $(A, B)$ to $J$ would lead to a relaxation which is computationally infeasible to solve.

**Proposition 2.1.** *The following two operations do not affect the set $\mathcal{L}(J)$, and thus relaxation* (2):
- *pick edges* $(A, B), (B, C) \in J$, *add* $(A, C)$ *to* $J$. (4a)
- *pick edges* $(A, B), (A, C) \in J$ *with* $B \supset C$, *add* $(B, C)$ *to* $J$. (4b)

A proof is given in Appendix A. We denote $\bar{J}$ the closure of $J$ with respect to these operations; in other words, $\bar{J}$ is obtained from $J$ by applying operations (4) while possible. We have $\mathcal{L}(\bar{J}) = \mathcal{L}(J)$.

We mention that taking the closure will not cost us anything: each pass of our final Algorithm 3 will use at most one message operation per factor in $\mathcal{F}$. Using $\bar{J}$ will be quite important; for example, it will allow us to extend an ordering on nodes to an ordering on factors in a consistent way.

**Reparameterization and dual problem** For each $(A, B) \in J$ let $m_{AB} = (m_{AB}(\boldsymbol{x}_B))$ be a *message* from $A$ to $B$. Each message vector $m = (m_{AB})$ defines a new vector $\theta = \bar{\theta}[m]$ according

to
$$\theta_B(\boldsymbol{x}_B) = \bar{\theta}_B(\boldsymbol{x}_B) + \sum_{A|(A,B)\in J} m_{AB}(\boldsymbol{x}_B) - \sum_{C|(B,C)\in J} m_{BC}(\boldsymbol{x}_C) \tag{5}$$

It is easy to check that $\bar\theta$ and $\theta$ define the same objective function, i.e. $f(\boldsymbol{x}\mid\bar\theta)=f(\boldsymbol{x}\mid\theta)$ for all labelings $\boldsymbol{x}\in\mathcal{X}$. Thus, $\theta$ is a *reparameterization* of $\bar\theta$ [34]. If $\theta=\bar\theta[m]$ for some vector $m$ then we will write this as $\theta\equiv\bar\theta$.

Using the notion of reparameterization, we can write the dual of (2) as follows [36]:
$$\max_{\theta\equiv\bar\theta}\sum_{A\in\mathcal{F}}\min_{\boldsymbol{x}_A}\theta_A(\boldsymbol{x}_A) \tag{6}$$

**Convex combination of subproblems** Let $\mathcal{T}$ be a set of subproblem indexes and $\rho:\mathcal{T}\to(0,1]$ be a probability distribution on $\mathcal{T}$ with $\sum_T \rho^T=1$. Each subproblem $T\in\mathcal{T}$ is characterized by the set of factors $\mathcal{F}_T\subseteq\mathcal{F}$. For factor $A\in\mathcal{F}$ let $\mathcal{T}_A=\{T\in\mathcal{T}\mid A\in\mathcal{F}_T\}$ be the set of subproblems containing $A$. For each $T\in\mathcal{T}$ we will have vector $\theta^T$ of the same dimension as $\bar\theta$. The collection of vectors $\theta^T$ will be denoted as $\boldsymbol{\theta}=(\theta^T\mid T\in\mathcal{T})$. Let $\Omega$ be the following constraint set for $\boldsymbol{\theta}$:

$$\Omega=\left\{\boldsymbol{\theta}\;\middle|\;\begin{array}{l}\theta_A^T(\boldsymbol{x}_A)=0\quad\forall T,A\in\mathcal{F}-\mathcal{F}_T,\boldsymbol{x}_A\\ \sum_T\rho^T\theta^T\equiv\bar\theta\end{array}\right\} \tag{7}$$

The first condition says that $\theta^T$ must respect the structure of subproblem $T$, while the second condition means that $\boldsymbol{\theta}$ is a $\rho$-*reparameterization* of $\bar\theta$ [34].

For a vector $\boldsymbol{\theta}=(\theta^T\mid T\in\mathcal{T})$ let us define
$$\Phi(\boldsymbol{\theta})=\sum_T\rho^T\min_{\boldsymbol{x}}f(\boldsymbol{x}\mid\theta^T) \tag{8}$$

Clearly, if $\boldsymbol{\theta}\in\Omega$ then $\Phi(\boldsymbol{\theta})$ is a lower bound on the minimum of function $f(\boldsymbol{x}\mid\bar\theta)$. Our goal will be to compute vector $\boldsymbol{\theta}\in\Omega$ that maximizes this bound, i.e. solve the problem
$$\max_{\boldsymbol{\theta}\in\Omega}\Phi(\boldsymbol{\theta}) \tag{9}$$

**Decomposition into junction trees** For a factor $A\in\mathcal{F}$ we denote $\mathcal{F}_A=\{B\in\mathcal{F}\mid(A,B)\in\bar{J}\}\cup\{A\}$. We say that factor $A\in\mathcal{F}$ is *outer* if it has no incoming edges in $(\mathcal{F},J)$ (or equivalently in $(\mathcal{F},\bar{J})$). The set of outer factors will be denoted as $\mathcal{O}\subseteq\mathcal{F}$. Non-outer factors will be called *separators*, and their set will be denoted as $\mathcal{S}=\mathcal{F}-\mathcal{O}$. Finally, for subproblem $T\in\mathcal{T}$ we denote $\mathcal{O}_T=\mathcal{O}\cap\mathcal{F}_T$.

In this paper we will be interested in decompositions satisfying the following properties:

1. *There holds $\mathcal{F}_T=\bigcup_{A\in\mathcal{O}_T}\mathcal{F}_A$. Thus, subproblem $T$ is completely specified by its set of outer factors $\mathcal{O}_T$.*
2. *There exists a junction tree $(\mathcal{O}_T,\mathcal{E}_T)$, i.e. a tree-structured graph $(\mathcal{O}_T,\mathcal{E}_T)$ with the running intersection property [3]: for any $A,B\in\mathcal{O}_T$ all factors $C\in\mathcal{O}_T$ on the unique path connecting $A$ and $B$ satisfy $A\cap B\subseteq C$.*
3. *For each $(A,B)\in\mathcal{E}_T$ there holds $A\cap B\in\mathcal{F}_A$ and $A\cap B\in\mathcal{F}_B$.*

In general, conditions $A,B\in\mathcal{F}$, $B\subseteq A$ don't imply $B\in\mathcal{F}_A$. However, the following holds (see Appendix B):

**Proposition 2.2.** *If $A,B\in\mathcal{F}_T$ and $B\subseteq A$ then $B\in\mathcal{F}_A$.*

We will restrict slightly allowed sets $J$ by assuming

4. *If $v\in A\in\mathcal{F}$ then $\{v\}\in\mathcal{F}_A$.*

and also allow only one tree per outer factor:

5. *There holds $|\mathcal{T}_A|=1$ for each $A\in\mathcal{O}$.*

The last condition is not really an inherent limitation[1], but it will help to simplify the presentation of the algorithm. Furthermore, in practice there is no clear reason to cover outer factors more than once.

---
[1] If we have a decomposition in which factor $A\in\mathcal{O}$ belongs to several trees, then we can do the following transformation: add to $V$ new "dummy" nodes $v_T$ for each $T\in\mathcal{T}_A$, add to $\mathcal{F}$ new outer factors $A\cup\{v_T\}$ with zero cost functions, add to $J$ edges $(A\cup\{v_T\},A)$, and finally assign $A\cup\{v_T\}$ to tree $T$.

# 3 TRW-S algorithm

We will start with a general version of the algorithm for an arbitrary decomposition into junction trees, and then present a more specialized version for *monotonic chains*.

We will need the following notation. For tree $T$ and factor $A \in \mathcal{F}_T$ we denote

$$\nu_A^T(\boldsymbol{x}_A) = \sum_{B \in \mathcal{F}_A} \theta_B^T(\boldsymbol{x}_B) \tag{10}$$

We say that $\nu_A^T$ gives *correct min-marginals for $T$* if

$$\nu_A^T(\boldsymbol{x}_A) = \min_{\boldsymbol{x}_{V-A}} f(\boldsymbol{x} \mid \theta_A^T) \quad \forall \boldsymbol{x}_A \tag{11}$$

## 3.1 General version of TRW-S

The algorithm will rely on two operations:

**1**. Average factor $B \in \mathcal{S}$:
- compute $\nu_B = \left(\sum_{T \in \mathcal{T}_B} \rho^T \nu_B^T\right) / \left(\sum_{T \in \mathcal{T}_B} \rho^T\right)$ (12a)
- update parameters $\theta_B^T$ for $T \in \mathcal{T}_B$ so that we get $\nu_B^T = \nu_B$ for all $T \in \mathcal{T}_B$ (12b)

**2**. Send message $A \to B$ in $T$ where $A, B \in \mathcal{F}_T$, $(A, B) \in \bar{J}$:
- compute $\delta^T(\boldsymbol{x}_B) = \min_{\boldsymbol{x}_{A-B}} \nu_A^T(\boldsymbol{x}_A) - \nu_B^T(\boldsymbol{x}_B) \quad \forall \boldsymbol{x}_B$ (13a)
- update $\theta_A^T(\boldsymbol{x}_A) := \theta_A^T(\boldsymbol{x}_A) - \delta^T(\boldsymbol{x}_B) \quad \forall \boldsymbol{x}_A$ and $\theta_B^T(\boldsymbol{x}_B) := \theta_B^T(\boldsymbol{x}_B) + \delta^T(\boldsymbol{x}_B) \quad \forall \boldsymbol{x}_B$ (13b)

Note that after update (13) message $A \to B$ becomes *valid* in $T$, i.e. there holds $\min_{\boldsymbol{x}_{A-B}} \nu_A^T(\boldsymbol{x}_A) = \nu_B^T(\boldsymbol{x}_B)$ for all $\boldsymbol{x}_B$. This is equivalent to

$$\min_{\boldsymbol{x}_{A-B}} \sum_{C \in \mathcal{F}_A - \mathcal{F}_B} \theta_C^T(\boldsymbol{x}_C) = 0 \quad \forall \boldsymbol{x}_B \tag{14}$$

The TRW-S algorithm simply performs min-marginal averaging operations for factors $B \in \mathcal{S}$:

---
**Algorithm 1** TRW-S
---
0: initialize $\boldsymbol{\theta}$ with some vector in $\Omega$
1: **repeat** until some stopping criterion
2:   **for** factors $B \in \mathcal{S}$ **do** in some fixed order that visits each factor in $\mathcal{S}$ at least once
3:     for each $T \in \mathcal{T}_B$ reparameterize $\theta^T$ so that $\nu_B^T$ gives correct min-marginals for $B$ (eq. 11)
4:     average $B$ using eq. (12)
5:   **end for**
6: **end repeat**
---

Note, Algorithm 1 is a special case of *tree-consistency bound optimization* (TBCO) from [16]. We postpone the analysis of this algorithm until section 4. One of the properties is the monotonic behaviour of the lower bound: $\Phi(\boldsymbol{\theta})$ never goes down. We also formally prove that the algorithm is characterized by the same stopping condition as the the min-sum diffusion algorithm [36] (up to reparameterization).

Step 3 of the algorithm requires computing min-marginals for factor $B$ in tree $T \in \mathcal{T}_B$. This can be done via a junction tree algorithm [3] in two steps as follows. (i) Choose a factor $A \in \mathcal{O}_T$ that contains $B$; make $A$ the root of tree $(\mathcal{O}_T, \mathcal{E}_T)$. For each directed edge $(C, D) \in \mathcal{E}_T$ oriented toward $A$ send a message $C \to S$ using eq. (13) where $S = C \cap D$. Do it in the "inward order" that starts from the leaves. (ii) If $A \neq B$ send a message $A \to B$ using (13).

It is not difficult to see that after step (i) $\nu_A^T$ gives correct min-marginals for $T$. A sketch of the proof is as follows. After sending message $C \to S$ from a leaf $C$ this message becomes valid, i.e. (14) holds. This means that removing factors $\{E \mid E \cap (C - S) \neq \varnothing\}$ from $\mathcal{F}_T$ will not affect min-marginals for the remaining factors. Applying this argument inductively gives the claim.

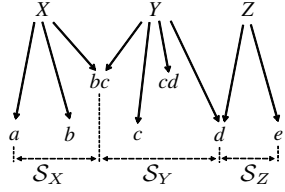

Figure 1: Example of a chain with three outer factors $X = abc$, $Y = bcd$, $Z = de$. (For brevity, factors $\{x, y, \ldots, z\}$ are written as $xy\ldots z$.) The order of factors in $\mathcal{S}$ is reflected by their $x$-coordinates.

## 3.2 TRW-S with monotonic chains

Running the junction tree algorithm from scratch every time would be very inefficient if trees are large. Fortunately, we can speed up computations by reusing previously passed messages. The general idea of not recomputing messages when they would not change has appeared several times in the literature in different contexts, e.g. in [7, 18, 9]. To make most of this idea, we now impose the following assumption on the decomposition; it will allow computing min-marginals by sending messages only from immediate neighbors.

6. *Each tree $(\mathcal{O}_T, \mathcal{E}_T)$ is a monotonic chain w.r.t. to some fixed total order $\leq$ on $V$, i.e. it is an ordered sequence of factors $A_1, \ldots, A_k$ such that for each pair of consecutive factors $(A_i, A_{i+1}) \in \mathcal{E}_T$ intersecting at $S = A_i \cap A_{i+1} \in \mathcal{S}$ there holds*

$$u < v < w \qquad \forall u \in A_i - S, v \in S, w \in A_{i+1} - S \tag{15}$$

The total order on factors in $\mathcal{O}_T$ corresponding to chain $T$ will be denoted as $\preceq^T$. From now on we will treat $\mathcal{E}_T$ as a *directed* set of edges that contains pairs $(A, A')$ with $A \prec^T A'$. It is convenient to define for factor $A \in \mathcal{O}_T$ "left" and "right" separators as

$$\texttt{sep}^- A = \begin{cases} A' \cap A & \text{if } \exists (A', A) \in \mathcal{E}_T \\ \{\min A\} & \text{if } A \text{ is the first factor in } T \end{cases} \tag{16a}$$

$$\texttt{sep}^+ A = \begin{cases} A \cap A' & \text{if } \exists (A, A') \in \mathcal{E}_T \\ \{\max A\} & \text{if } A \text{ is the last factor in } T \end{cases} \tag{16b}$$

Here $\min$ and $\max$ are taken w.r.t. to $\leq$; therefore, $\{\min A\}$ and $\{\max A\}$ are singleton separators in $\mathcal{F}_A$. Note, we dropped the dependence of $\texttt{sep}^- A, \texttt{sep}^+ A$ on $T$ due to Assumption 5.

**Algorithm** First, we select an ordering $\preceq$ on $\mathcal{S}$ that *extends* ordering $\leq$, i.e. the following holds:
• *if $\min A < \min B$ and $\max A \leq \max B$ then $A \prec B$;*
• *if $\max A > \max B$ and $\min A \geq \min B$ then $A \succ B$.*
This can be done in several ways, e.g. by choosing a unique sequence $\sigma_A = (\min A, \max A, \ldots)$ for each $A \in \mathcal{S}$ and then setting $\preceq$ as the lexicographical order on $\sigma_A$ (using $\leq$ for comparing components of $\sigma_A$).

The choice of $\preceq$ will determine the order of averaging operations: the algorithm will alternate between a forward pass (processing factors in $\mathcal{S}$ in the order $\preceq$), and a backward passes (which uses the reverse order).

For a factor $A \in \mathcal{O}$ we define (see Fig. 1)

$$\mathcal{S}_A = \{B \in \mathcal{F}_A \cap \mathcal{S} \mid \texttt{sep}^- A \preceq B \preceq \texttt{sep}^+ A\} \tag{17}$$

It is possible to prove the following (see Appendix C):

**Proposition 3.1.** *If ordering $\preceq$ extends $\leq$ and $T$ is a monotonic chain w.r.t. $\leq$ then $\mathcal{F}_T \cap \mathcal{S} = \bigcup_{A \in \mathcal{O}_T} \mathcal{S}_A$.*

We now formulate the TRW-S algorithm.

**Remark 1** It follows from Proposition 3.1 and definition (17) that in step 3 there exists exactly one factor $A$ with stated properties, with one exception: if $B$ is the first factor in $\mathcal{F}_T \cap \mathcal{S}$ (i.e. $B = \texttt{sep}^- A_1$ where $A_1 \in \mathcal{O}_T$ is the first factor of chain $T$) then no such $A$ exists. Note, we do not send messages $A \to \texttt{sep}^- A$ for $A \in \mathcal{O}$ since these messages remain valid from the previous reverse pass (see the analysis in section 4).

**Remark 2** As we will show later, sometimes we may speed up message passing operations. Consider the example in Fig. 1. When passing message $Y \to c$ in the forward pass, we know that message $Y \to bc$ is valid (from the previous reverse pass); therefore, we can compute increment $\delta^T(\boldsymbol{x}_c)$ in (13a) by going through labelings $\boldsymbol{x}_{bc}$ rather than through labelings $\boldsymbol{x}_Y$.

**Algorithm 2** TRW-S with monotonic chains

0: initialize $\boldsymbol{\theta} \in \Omega$
1: **for each** $B \in \mathcal{S}$ **do** in the order $\preceq$
2: | **for each** $T \in \mathcal{T}_B$ **do**
3: | | find $A \in \mathcal{O}_T$ with $B \in \mathcal{S}_A$, $B \neq \mathtt{sep}^- A$; if it exists, send message $A \to B$ in $T$ (eq. 13)
4: | **end for**
5: | average $B$ using (12)
6: **end for**
7: if a stopping criterion is satisfied, terminate; otherwise reverse the ordering and go to step 1

---

Now consider message $X \to b$ in the forward pass. Message $X \to bc$ is invalid at this point, so we cannot use the trick above. However, we can instead "preemptively" compute message $X \to bc$ (without reparameterizing anything), and then use it both for $b$ and $bc$. Details are given in the next section.

### 3.3 Implementation via messages

It is easy to see that each step of Algorithm 2 preserves property $\theta_C^T = \theta_C^{T'}$ for $T, T' \in \mathcal{T}_C$, $C \in \mathcal{F}$ (assuming that it holds after initialization). Therefore, it suffices to store the *cumulative* vector $\theta = \sum_T \rho^T \theta^T$; components of vector $\boldsymbol{\theta} = (\theta^T)$ are then given by

$$\theta_C^T = \frac{1}{\rho_C} \theta_C \qquad \forall C \in \mathcal{F}, T \in \mathcal{T}_C \tag{18}$$

where $\rho_C = \sum_{T \in \mathcal{T}_C} \rho^T$ is the *factor appearance probability*. By construction, vector $\theta$ is a reparameterization of $\bar{\theta}$ (eq. 5), so we can store it via messages $m = (m_{AB} \mid (A, B) \in J)$ where $J = \{(A, B) \mid A \in \mathcal{O}, B \in \mathcal{S}_A\}$. We thus have

$$\theta_A(\boldsymbol{x}_A) = \bar{\theta}_A(\boldsymbol{x}_A) - \sum_{(A,B) \in J} m_{AB}(\boldsymbol{x}_B) \qquad \forall A \in \mathcal{O} \tag{19a}$$

$$\theta_B(\boldsymbol{x}_B) = \bar{\theta}_B(\boldsymbol{x}_B) + \sum_{(A,B) \in J} m_{AB}(\boldsymbol{x}_B) \qquad \forall B \in \mathcal{S} \tag{19b}$$

For efficiency reasons we will also store vectors $\theta_B$ for $B \in \mathcal{S}$ explicitly, so that we don't need to recompute them from $m$ every time. The resulting algorithm is given below.

**Algorithm 3** TRW-S with monotonic chains

0: set $m_{AB} := \mathbf{0} \quad \forall (A, B) \in J$ and $\theta_B := \bar{\theta}_B \quad \forall B \in \mathcal{S}$
1: **for each** $B \in \mathcal{S}$ **do** in the order $\preceq$
2: | set $\theta_B := \bar{\theta}_B$
3: | **for each** $(A, B) \in J$ **do**
4: | | **if** $B \neq \mathtt{sep}^- A$ **then**
5: | | | update

$$m_{AB}(\boldsymbol{x}_B) := \min_{\boldsymbol{x}_{A-B}} \left[ \bar{\theta}_A - \sum_{\substack{(A,C) \in J \\ C \neq B}} m_{AC}(\boldsymbol{x}_C) + \sum_{C \in \mathcal{F}_A \cap \mathcal{S} - \mathcal{F}_B} \frac{\rho_A}{\rho_C} \theta_C(\boldsymbol{x}_C) \right] \tag{20}$$

6: | | | compute $\gamma = \min_{\boldsymbol{x}_B} m_{AB}(\boldsymbol{x}_B)$, update $m_{AB}(\boldsymbol{x}_B) -\!= \gamma$ /* optional: for numerical stability */
7: | | **end if**
8: | | update $\theta_B +\!= m_{AB}$
9: | **end for**
10: **end for**
11: if a stopping criterion is satisfied, terminate; otherwise reverse the ordering and go to step 1

---

**Reusing messages in nested factors** Suppose that we have two factors $P, B \in \mathcal{S}_A$, $A \in \mathcal{O}$ with $B \subset P$ such that $B$ is processed immediately **after** $P$ in chain $T \in \mathcal{T}_A$, i.e. there are no other factors in $\mathcal{S}_A$ between $P$ and $B$. When processing edge $(A, B)$, we know that $(A, P)$ contains a valid message. This allows us to speed up the computation of message from $A$ to $B$. Namely, we need to perform the update $m_{AB} += \rho_A \delta^T$ where $\delta^T(\bm{x}_B) = \min_{\bm{x}_{A-B}} \nu_A^T(\bm{x}_A) - \nu_B^T(\bm{x}_B) = \min_{\bm{x}_{P-B}} \nu_P^T(\bm{x}_P) - \nu_B^T(\bm{x}_B)$. Thus, the update in step 5 can be replaced by the equivalent update

$$m_{AB}(\bm{x}_B) += \min_{\bm{x}_{P-B}} \sum_{C \in \mathcal{F}_P - \mathcal{F}_B} \frac{\rho_A}{\rho_C} \theta_C(\bm{x}_C)$$

Now suppose that $P, B \in \mathcal{S}_A$, $A \in \mathcal{O}$, $B \subset P$ and $B$ is processed immediately **before** $P$, i.e. there are no other factors in $\mathcal{S}_A$ between $B$ and $P$. In that case we can replace step 5 for factor $B$ with the following:

(a) set $m_{AP}^\circ := m_{AP}$, update $m_{AP}$ as in step 5 (where $B$ is replaced with $P$), set $\delta_{AP} := m_{AP} - m_{AP}^\circ$

(b) compute

$$\delta(\bm{x}_B) := \min_{\bm{x}_{P-B}} \left[ \delta_{AP}(\bm{x}_P) + \sum_{C \in \mathcal{F}_P - \mathcal{F}_B} \frac{\rho_A}{\rho_C} \theta_C(\bm{x}_P) \right]$$

(c) update $m_{AB} += \delta$ and $m_{AP}(\bm{x}_P) -= \delta(\bm{x}_B)$

It can be checked that (i) the resulting message $m_{AB}$ is the same as the one that would be computed in step 5; (ii) when passing message $A \to P$ (during the averaging step for $P$), the update in step 5 would not change $m_{AP}$. Thus, the latter update can be skipped (though the normalization step 6 still needs to be applied). Note, in operations (a)-(c) we modify $m_{AP}$ but do not change $\theta_P$, therefore equality (19b) for factor $P$ temporarily becomes violated (but gets restored after processing $P$).

## 4 Algorithm's analysis

We will first analysis the general version of TRW-S (Algorithm 1). We will then show that after the first forward pass Algorithm 2 is a special case of Algorithm 1: during the averaging step 5 vectors $\nu_B^T$ give correct min-marginals for trees $T \in \mathcal{T}_B$.

### 4.1 Analysis of Algorithm 1

We will need a few definitions. Consider subset $A \subseteq V$ and a vector $\varphi_A$ with components $\varphi_A(\bm{x}_A)$. We define relation $\langle \varphi_A \rangle \subseteq \otimes_{v \in A} \mathcal{X}_v$ as

$$\langle \varphi_A \rangle = \{ \bm{x}_A \mid \varphi_A(\bm{x}_A) = \min_{\bm{x}_A'} \varphi_A(\bm{x}_A') \} \tag{21}$$

For a tree $T \in \mathcal{T}$ we define vector $\nu^T$ with components $(\nu^T(\bm{x}) \mid \bm{x} \in \mathcal{X})$ via

$$\nu^T(\bm{x}) = f(\bm{x} \mid \theta^T) = \sum_{B \in \mathcal{F}_T} \theta_B^T(\bm{x}_B) \tag{22}$$

This can be viewed as a generalization of definition (10). We emphasize that vectors $\nu^T$ and $\nu_A^T$ for $A \in \mathcal{F}_T$ are uniquely determined by vector $\theta^T$ via a linear transformation.

A *projection* of relation $\mathcal{R} \subseteq \otimes_{v \in A} \mathcal{X}_v$ to subset $B \subseteq A$ is defined as

$$\pi_B(\mathcal{R}) = \{ \bm{x}_B \mid \bm{x}_A \in \mathcal{R} \} \tag{23}$$

(Recall that $\bm{x}_B$ is the restriction of labeling $\bm{x}_A$ to $B$).

**Weak tree agreement** We now define a condition characterizing a stopping criterion for TRW-S.

**Definition 4.1.** *Vector $\bm{\theta} = (\theta^T \mid T \in \mathcal{T})$ is said to satisfy the* enhanced weak tree agreement *(EWTA) condition for factor $B \in \mathcal{F}$ if $\pi_B(\langle \nu^T \rangle) = \pi_B(\langle \nu^{T'} \rangle)$ for $T, T' \in \mathcal{T}_B$.*

*It satisfies the* weak tree agreement *(WTA) for $B \in \mathcal{F}$ if there exist non-empty relations $(\mathcal{R}^T \subseteq \langle \nu^T \rangle \mid T \in \mathcal{T})$ s.t. $\pi_B(\mathcal{R}^T) = \pi_B(\mathcal{R}^{T'})$ for $T, T' \in \mathcal{T}_B$.*

*Vector $\bm{\theta}$ is said to satisfy EWTA (WTA) if it satisfies EWTA (WTA) for all $B \in \mathcal{F}$.*

Clearly, EWTA implies WTA (but not the other way around).

**Theorem 4.2.** *Let $\boldsymbol{\theta}$, $\tilde{\boldsymbol{\theta}}$ be respectively the vectors before and after averaging step 4 for factor $B$.*
*(a) The lower bound does not decrease: $\Phi(\tilde{\boldsymbol{\theta}}) \geq \Phi(\boldsymbol{\theta})$.*
*(b) If $\boldsymbol{\theta}$ satisfies WTA for $B$ with relations $(\mathcal{R}^T \mid T \in \mathcal{T})$, then $\tilde{\boldsymbol{\theta}}$ also satisfies WTA with the same set of relations. Furthermore, $\Phi(\tilde{\boldsymbol{\theta}}) = \Phi(\boldsymbol{\theta})$.*
*(c) If $\Phi(\tilde{\boldsymbol{\theta}}) = \Phi(\boldsymbol{\theta})$ then $\langle \tilde{\nu}^T \rangle \subseteq \langle \nu^T \rangle$ for each $T \in \mathcal{T}$. (d) If $\Phi(\tilde{\boldsymbol{\theta}}) = \Phi(\boldsymbol{\theta})$ and $\boldsymbol{\theta}$ does not satisfy EWTA for $B$ then $\langle \tilde{\nu}^T \rangle \subset \langle \nu^T \rangle$ for at least one tree $T \in \mathcal{T}_B$.*

A proof is given in Appendix D.

**Corollary 4.3.**
- *If $\boldsymbol{\theta}$ satisfies WTA then Algorithm 1 will not increase the lower bound $\Phi(\boldsymbol{\theta})$, and furthermore after a finite number of steps $\boldsymbol{\theta}$ will satisfy EWTA.*
- *If $\boldsymbol{\theta}$ does not satisfy WTA then bound $\Phi(\boldsymbol{\theta})$ will increase after a finite number of steps.*

*Proof.* The first claim follows from parts (b,d) of theorem 4.2. To prove the second claim, assume that $\Phi(\boldsymbol{\theta})$ stays constant after an arbitrary number of steps. From parts (c,d) we conclude that after a finite number of steps we get vector $\tilde{\boldsymbol{\theta}}$ satisfying EWTA such that $\langle \tilde{\nu}^T \rangle \subseteq \langle \nu^T \rangle$ for all $T$. This means that $\boldsymbol{\theta}$ satisfies WTA with relations $\mathcal{R}^T = \langle \tilde{\nu}^T \rangle$. □

**Relation to min-sum diffusion** We now show that WTA condition is closely related to the stopping criterion of the MSD algorithm [36]. Recall that MSD tries to maximize lower bound

$$\Psi(\theta) = \sum_{A \in \mathcal{F}} \min_{\boldsymbol{x}_A} \theta_A(\boldsymbol{x}_A) \qquad (24)$$

over vectors $\theta \equiv \bar{\theta}$. Its stopping criterion is described in the following definition.

**Definition 4.4.** *Vector $\theta \equiv \bar{\theta}$ is said to satisfy the* enhanced $J$-consistency *condition if $\pi_B(\langle \theta_A \rangle) = \langle \theta_B \rangle$ for each $(A, B) \in J$. It is said to satisfy the $J$-consistency condition if there exist non-empty relations $(\mathcal{R}_B \subseteq \langle \theta_B \rangle \mid B \in \mathcal{F})$ such that $\pi_B(\mathcal{R}_A) = \mathcal{R}_B$ for each $(A, B) \in J$.*

We denote $\Omega^*$ to be set of vectors $\boldsymbol{\theta} \in \Omega$ that satisfy the WTA condition, and $\Lambda^*$ to be the set of vectors $\theta \equiv \bar{\theta}$ that satisfy the $J$-consistency condition.

**Theorem 4.5.** *There exist mappings $\phi : \Omega^* \to \Lambda^*$ and $\psi : \Lambda^* \to \Omega^*$ that preserve the value of the lower bound, i.e. $\Psi(\phi(\boldsymbol{\theta})) = \Phi(\boldsymbol{\theta})$ and $\Phi(\psi(\theta)) = \Psi(\theta)$.*

A proof is given in Appendix E.

### 4.2 Analysis of Algorithm 2

We now analyze the TRW-S algorithm with monotonic chains. In order to do this, we will reformulate it slightly. Namely, we will maintain factor $\text{CUR}_T \in \mathcal{O}_T$ for each $T \in \mathcal{T}$ ("current outer factor of chain $T$") and factor $\text{CHILD}_A \in \mathcal{S}_A$ for each $A \in \mathcal{O}$:

---

0: initialize $\boldsymbol{\theta} \in \Omega$
   for each $T \in \mathcal{T}$ set $\text{CUR}_T$ = first factor of chain $T$
   for each $A \in \mathcal{O}$ set $\text{CHILD}_A = \text{sep}^- A$
1: **for each** $B \in \mathcal{S}$ **do** in the order $\preceq$
2: | **for each** $T \in \mathcal{T}_B$ **do**
3: | | let $A = \text{CUR}_T$
4: | | if $\text{CHILD}_A \neq B$ then send message $A \to B$ in $T$ (eq. 13) and update $\text{CHILD}_A := B$
5: | | if $B = \text{sep}^+ A$ and $\exists (A, A') \in \mathcal{E}_T$ set $\text{CUR}_T := A'$
6: | **end for**
7: | average $B$ using (12)
8: **end for**
9: if a stopping criterion is satisfied, terminate; otherwise reverse the ordering and go to step 1

---

It should be clear that this algorithm is equivalent to Algorithm 2. In particular, the following is maintained:

**Proposition 4.6.** *(a) In step 4 there holds $B \in \mathcal{S}_A$.*
*(b) If $A' \in \mathcal{O}_T$, $A' \prec^T \mathtt{CUR}_T$ then $\mathtt{CHILD}_{A'} = \mathtt{sep}^+ A'$.*
*(c) If $A' \in \mathcal{O}_T$, $A' \succ^T \mathtt{CUR}_T$ then $\mathtt{CHILD}_{A'} = \mathtt{sep}^- A'$.*

The algorithm's correctness will follow from

**Theorem 4.7.** *(a) Each step of the algorithm preserves the validity of edges $(A, \mathtt{CHILD}_A)$, $A \in \mathcal{O}_T$ in $T$: if the edge contained a valid message in $T$ before the step (eq. 14), then this message remains valid afterwards.*
*(b) After the first forward pass, all edges $(A, \mathtt{CHILD}_A)$, $A \in \mathcal{O}_T$ are valid in $T$. Consequently, in step 5 vector $\nu_B^T$ gives correct min-marginals in $T$ for each $T \in \mathcal{T}_B$.*

*Proof.* Consider loop 1-8 for factor $B$, and let us fix tree $T \in \mathcal{T}_B$. Let $A$ be the factor defined in step 3: $A = \mathtt{CUR}_T$. It is clear that sending message $A \to B$ in $T$ makes edge $(A, B)$ valid, and that averaging $B$ in step 7 preserves the validity of this edge (see eq. 14).

Now consider factor $A' \in \mathcal{O}_T$, $A' \prec A$, and define $S = \mathtt{CHILD}_{A'} = \mathtt{sep}^+ A'$. Let us show that update of vectors $\theta_C^T$ for $C \in \mathcal{F}_A$ preserves the validity of edge $(A', S)$ in $T$. We need to prove that $C \notin \mathcal{F}_{A'} - \mathcal{F}_S$ (since the definition of a valid edge involves only vectors $\theta_D^T$ for $D \in \mathcal{F}_{A'} - \mathcal{F}_S$). Suppose that $C \in \mathcal{F}_{A'}$. By the running intersection property we have $C \subseteq A''$ where $A''$ is the right neighbor of $A'$, i.e. $(A', A'') \in \mathcal{E}_T$. Therefore, $C \subseteq A' \cap A'' = \mathtt{sep}^+ A' = S$, and so $C \in \mathcal{F}_S$ and $C \notin \mathcal{F}_{A'} - \mathcal{F}_S$, as claimed.

A similar argument can be used for factors $A' \in \mathcal{O}_T$, $A' \succ A$. Part (a) is proved. Part (b) easily follows from part (a) and the fact that step 4 makes edge $A \to B$ valid in $T$. □

## 5 Experimental results

We compare the proposed TRW-S to (our own implementations[2] of) min-sum diffusion (MSD) [36], MPLP [27] and subgradient ascent methods (SG) [11], the latter with (non-monotonic) chains where each outer factor belongs to exactly one chain[3]. Our current implementation of TRW-S does not support the second "reuse" scheme described in the end section 3.3. Since timings are implementation-dependent we also report a "message effort measure", where each minimization computation over a factor of size $n$ contributes $n$. All experiments were run on a Core i5 machine with 2.5 GHz.

We evaluate the methods on problems from the fields of computer vision and natural language processing: we consider image segmentation with a generalized Potts model with 2x2 blocks, with factor-based curvature, with constraint-based curvature and with histogram-based data terms. Also, we consider stereo disparity estimation with second order differences, and word alignment. For stereo there are 8 labels per variable, for the generalized Potts model 4 and for all other problems 2.

Three of our problems (2x2 block Potts, stereo and factor-based curvature) use factors of low order only, so they are explored with singleton and pairwise separators (same style of message computation subroutines for all compared schemes). The remaining problems are of high order (16, 9600 and 5281 resp.). Constraint-based curvature requires handling integer linear constraints, where we use the method of [19] for the message computations. Histogram image segmentation and word alignment require cardinality potentials, the latter also uses 1-of-N potentials. We handle this as in [31] and implemented specilazied routines for MSD with these high order terms. Here, MPLP has an advantage over TRW-S: with the specialized computations it effectively only needs to visit

---
[2]The code at http://cs.nyu.edu/~dsontag/code/ only supports factors up to size three.

[3] We used the step size rule that resembles the one in [13], namely $\lambda/(K+1)$ where $K$ is the number of times an iteration produced an inferior bound. We tried several $\lambda$s and chose the one that performs best after 500 iterations (for a given instance). We also tested the step-size rule from [12] for problems in the top row of Table 1, but it was inferior to our rule. A potential reason is that the rule of [12] depends on the the primal integral solution, and so if e.g. the relaxation is not tight then the gap will always remain large. (Note, in this case the step size doesn't go to zero, so this rule doesn't guarantee convergence to the optimum.) We mention that for stereo TRW-S was pretty close to the optimum after 250 iters, while the primal solution of SG was still far after 500 iters.

We also informally tested the step-size rule from [32], but found it to be inferior as well.

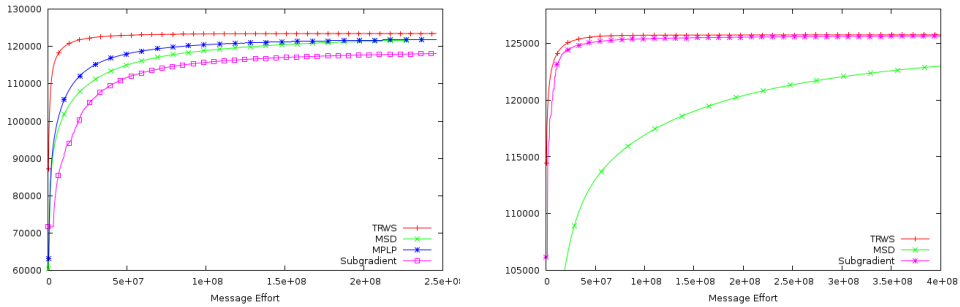

Figure 2: Plots of energy vs. message effort for singleton (left) and pair separators (right) for stereo.

|  | Gen. Potts | | | | Factor Curvature | | | | 2nd Order Stereo | | | |
|---|---|---|---|---|---|---|---|---|---|---|---|---|
|  | Bound | Time | MEff | Mem | Bound | Time | MEff | Mem | Bound | Time | MEff | Mem |
| MPLP | 7734036 | 39 | 115M | 42M | 22858924 | 256 | 734M | 51M | 121841.2 | 125 | 246M | 27M |
| MSD | 7731966 | 43 | 115M | 42M | 22884886 | 253 | 734M | 51M | 121862.7 | 131 | 246M | 27M |
| TRW-S | 7737053* | 25 | 86M | 45M | 22893060 | 78 | 532M | 71M | 123421.0 | 34 | 163M | 37M |
| SG | 7680893 | 23 | 29M | 46M | 22672531 | 105 | 202M | 86M | 115180.4 | 54 | 81M | 44M |

|  | Constraint Curvature | | | | Histogram Segmentation | | | | Word Alignment | | | |
|---|---|---|---|---|---|---|---|---|---|---|---|---|
|  | Bound | Time | MEff | Mem | Bound | Time | MEff | Mem | Bound | Time | MEff | Mem |
| MPLP | 23640257 | 794 | 14G | 156M | 40881 | 35 | 47G | 11M | 7650 | 57 | 50G | 52M |
| MSD | 24189191 | 832 | 14G | 156M | 41714 | 190 | 47G | 11M | 8435 | 220 | 50G | 52M |
| TRW-S | 24209662 | 3971 | 12G | 243M | 41756 | 8375 | 47G | 14M | 8127 | 9749 | 50G | 62M |
| SG | 22487409 | 517 | 1.1G | 268M | 41807 | 62 | 28M | 16M | 6543 | 130 | 129M | 65M |

Table 1: Singleton separators: relaxation values, timings (in seconds), message effort and memory for the compared schemes. Timings *exclude* any time spent on computing the intermediate bounds. We ran 250 iterations of TRW-S (forward+backward passes) and 500 of all other methods. MPLP and MSD can probably be sped up at the cost of extra memory. A "*" indicates that the method converged before the set number of iterations was used up.

each factor once per iteration (as is always the case for the subgradient method). TRW-S needs to visit each factor multiple times per iteration, so it is much slower. However, immense speedups in TRW-S should be possible by using advanced data structures. Consider, for example, a cardinality-dependent factor $A$ with binary labels. Message computation requires sorting certain values for nodes $v \in A$. Each TRW-S update changes only one of these values, so we can use e.g. 2-3-4 trees for maintaining a sorted order. This gives $O(\log |A|)$ time per node, same as in the other techniques. We left it as a future work.

**Singleton Separators** Table 1 compares the four methods with singleton separators on all problems. For problems of low order TRW-S performs always best, using less message effort than MSD and MPLP. SG used up less message effort, but still has higher running times: handling and projecting the gradients takes time, and one also has to compute minimizers along with the minimal values. Figure 2 plots how the energies evolve w.r.t. message effort on stereo for the different methods. For the high order terms TRW-S is beaten once, for histogram segmentation and by SG. To get the running times competitive one will need to use advanced data structures.

**Pairwise Separators** Experiments with pair separators are evaluated in Table 2, as mentioned only for low-order problems. A plot for stereo is provided in Figure 2. Again, TRW-S is beaten once by the subgradient method, this time for factor-based curvature. Possibly a different variable order might boost TRW-S here. Otherwise TRW-S performs best. It always outperforms MSD and due to the reuse scheme each iteration is also faster. For problems with a large number of pair separators SG finally profits from its reduced message effort: for the Potts model it is clearly fastest after a comparable number of iterations.

|        | **Gen. Potts** | | | | **Factor Curvature** | | | | **2nd Order Stereo** | | | |
|--------|---------|------|------|-----|----------|------|------|------|----------|------|------|------|
|        | Bound   | Time | MEff | Mem | Bound    | Time | MEff | Mem  | Bound    | Time | MEff | Mem  |
| MSD    | 7736742 | 514  | 463M | 49M | 22904652 | 202  | 1.8G | 110M | 124183.7 | 951  | 682M | 230M |
| TRW-S  | 7737053* | 379 | 202M | 59M | 22903356 | 171  | 724M | 162M | 125725.9 | 248  | 273M | 265M |
| SG     | 7700442 | 114  | 58M  | 47M | 23220873 | 136  | 202M | 117M | 125454.1 | 423  | 136M | 227M |

Table 2: Pair separators: experiments with low-order factors. We give relaxation values, timings, message effort and memory consumption. A "*" indicates the same as above.

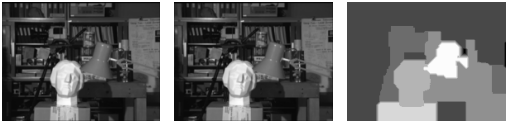

Figure 3: Data and results for second order stereo.

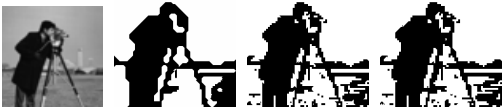

Figure 4: Data and results for curvature. From left to right: input image, result with constraint-based curvature, result with factor-based curvature and singleton separators and result with factor-based curvature and pair-separators.

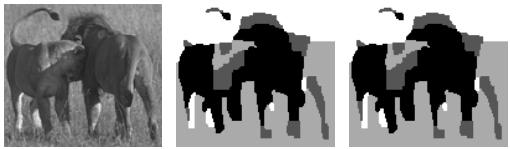

Figure 5: Data and results for the 2x2 Potts model. We show (near-identical) derived integral solutions with singleton and pair-separators.

### 5.1 Details on the Experiments

For second order stereo we use triplet factors in both horizontal and vertical direction. Each factor has the form

$$\theta_C(l_1, l_2, l_3) = \begin{cases} 0 & \text{if } |l_1 - l_2| \leq 1 \text{ and } |l_2 - l_3| \leq 1 \\ & \text{and } |(l_1 - l_2) - (l_2 - l_3)| = 0 \\ \lambda & \text{if } |l_1 - l_2| \leq 1 \text{ and } |l_2 - l_3| \leq 1 \\ & \text{and } |(l_1 - l_2) - (l_2 - l_3)| = 1 \\ 3\lambda & \text{else,} \end{cases}$$

with $\lambda = 15$. If there are only singleton separators we use a specialized message computation routine, otherwise a generic one. We run this on a downsized Tsukuba instance (half-scale, resulting in 8 disparities) shown in Figure 3.

For both factor-based [4, 28] and constraint-based curvature [24] we use an 8-connectivity with squared differences in the data term, a curvature weight of 10000 and no length weight. We apply this to a $64 \times 64$ pixel version of the cameraman image (Figure 4). Our implementation is based on RegionCurv[4].

The generalized Potts model is run on the lions image from Figure 5, where we use a block-weight of 5000.

---

[4] https://github.com/PetterS/regioncurv

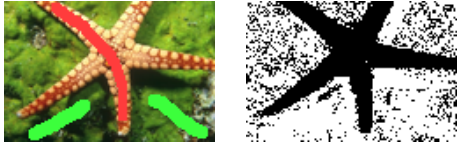

Figure 6: Data and results for histogram image segmentation. The integrality gap is large, but more refined strategies to obtain integral solutions are conceivable.

Histogram segmentation [33] is run on the sea star image in Figure 6, with the shown seed nodes and a prior weight of 2.

For word alignment [23] we use 100 sentences from the Italian-English Europarl corpus. Note that this problem has a much more irregular structure than the computer vision problems.

## 6 Conclusions

We showed how to generalize the TRW-S algorithm from pairwise MRFs to arbitrary graphical models. In order to improve efficiency, we had to overcome several challenges: (i) Find a suitable definition of monotonic junction chains that depends only on the order on nodes, and then extend this order to other factors in a consistent way. (ii) Make sure that parameters for the same factor in different chains stay the same (thus allowing an implementation via messages); we achieved this by passing messages only from outer factors. (iii) Find a way to reuse message computations in nested factors.

TRW-S has shown a good performance for pairwise graphical models [29, 30], and is among state-of-the-art techniques for problems such as stereo [22][5]. It has also been shown that tightening the relaxation by adding higher-order constraints (e.g. short cycles) is an effective strategy for solving challenging instances [25, 2]. Our work allows to combine the tightening strategy and the TRW-S technique; given results in [22, 25, 2], it is reasonable to assume that this would yield a state-of-the-art method for some applications.

In our experiments we pursued a different direction: applying generalized TRW-S directly to high-order graphical models. TRW-S outperformed MSD and MPLP on a number of applications. A notable exception is the word alignment problem where MSD was faster. At times the subgradient method beats TRW-S, but it is also often heavily inferior and requires the tuning of a step-size parameter.

Based on the above, we hope that generalized TRW-S will become one of the standard tools for MAP-MRF inference. Our implementation is available from [1].

One of the disadvantages of TRW-S is that it is not guaranteed to solve the LP: similarly to MSD and MPLP, it can get stuck in a suboptimal point. We see three ways to address this issue: (1) Interleave TRW-S and another technique that is guaranteed to solve the LP, e.g. a subgradient ascent. (2) Instead of squeezing the last bit from the current LP relaxation, one can tighten the relaxation by adding higher-order constraints as in [25, 2] and run TRW-S again. (3) Use a smoothed version of TRW-S [22]. At the moment such version has been presented only for the standard (pairwise) TRW-S, but we believe that generalizing it using our scheme should not be too difficult (we need to replace the max-product BP with the sum-product version).

Note that for some applications suboptimality of message passing techniques does not seem to be an issue: TRW either yields a global optimum or gets very close [37, 29].

---

[5]This holds for a CPU implementation, since the "oracle call" of smoothed TRW-S in [22] was 5-10 times slower than that of the simple TRW-S. On GPU the "oracle calls" took similar times, and so smoothed TRW-S outperformed simple TRW-S.

## Appendix A: proof of proposition 2.1

**Part (a)** For each $\boldsymbol{x}_C$ we can write

$$\sum_{\boldsymbol{x}_{A-C}} \mu_A(\boldsymbol{x}_A) = \sum_{\boldsymbol{x}_{B-C}} \sum_{\boldsymbol{x}_{A-B}} \mu_A(\boldsymbol{x}_A) \stackrel{(1)}{=} \sum_{\boldsymbol{x}_{B-C}} \mu_B(\boldsymbol{x}_B) \stackrel{(2)}{=} \mu_C(\boldsymbol{x}_C)$$

where (1) holds since $(A,B) \in J$ and (2) holds since $(B,C) \in J$. Thus, the constraint for $(A,C)$ follows from constraints for $(A,B), (B,C)$.

**Part (b)** For each $\boldsymbol{x}_C$ we can write

$$\sum_{\boldsymbol{x}_{B-C}} \mu_B(\boldsymbol{x}_B) \stackrel{(1)}{=} \sum_{\boldsymbol{x}_{B-C}} \sum_{\boldsymbol{x}_{A-B}} \mu_A(\boldsymbol{x}_A) = \sum_{\boldsymbol{x}_{A-C}} \mu_A(\boldsymbol{x}_A) \stackrel{(2)}{=} \mu_C(\boldsymbol{x}_C)$$

where (1) holds since $(A,B) \in J$ and (2) holds since $(A,C) \in J$. Thus, the constraint for $(B,C)$ follows from constraints for $(A,B), (A,C)$.

## Appendix B: proof of proposition 2.2

Since set $\bar{J}$ is closed under operations (4a)-(4b), we get

- If $B \in \mathcal{F}_A, C \in \mathcal{F}_B$ then $C \in \mathcal{F}_A$. (25a)
- If $B, C \in \mathcal{F}_A, B \supseteq C$ then $C \in \mathcal{F}_B$. (25b)

First, let us prove the proposition assuming that $A \in \mathcal{O}_T$. Consider $B \in \mathcal{F}_T$, $B \subset A$. Pick a factor $A' \in \mathcal{O}_T$ with $B \in \mathcal{F}_{A'}$ (it exists by Assumption 1); if there are several such factors, pick a one for which the distance from $A$ to $A'$ in the tree $(\mathcal{O}_T, \mathcal{E}_T)$ is minimal. We need to show that this distance is zero, i.e. $A = A'$. Suppose not; let $A'' \in \mathcal{O}_T$ be the neighbor of $A'$ (i.e. $(A', A'') \in \mathcal{E}_T$) which is closer to $A$ than $A'$. By the running intersection property, $A' \cap A \subseteq A''$, and so $B \subseteq A''$. Denote $S = A' \cap A''$; as we showed, $B \subseteq S$. By Assumption 3, $S \in \mathcal{F}_{A'}$ and $S \in \mathcal{F}_{A''}$. Since $B \in \mathcal{F}_{A'}$, we have $B \in \mathcal{F}_S$ by property (25b). Property (25a) and the fact $S \in \mathcal{F}_{A''}$ then gives $B \in \mathcal{F}_{A''}$. This contradicts to the choice of $A'$.

It remains to prove the proposition in the case when $A \in \mathcal{F}_T - \mathcal{O}_T$. Pick a factor $A' \in \mathcal{O}_T$ with $A \in \mathcal{F}_{A'}$. As we showed above, we have $B \in \mathcal{F}_{A'}$, therefore property (25b) gives $B \in \mathcal{F}_A$.

## Appendix C: proof of proposition 3.1

We need to show that for each $B \in \mathcal{F}_T \cap \mathcal{S}$ there exists $A \in \mathcal{O}_T$ with $B \in \mathcal{S}_A$. Assume that $|\mathcal{O}_T| \geq 2$ and thus $|A| \geq 2$ for all $A \in \mathcal{O}_T$, otherwise the claim is trivial.

Let $A_1, \ldots, A_k$ be the sequence of factors in $\mathcal{O}_T$. Denote $u_i = \min A_i$, $v_i = \max A_i$ (w.r.t. $\leq$). We have $u_i = \min \texttt{sep}^- A_i < \min \texttt{sep}^+ A_i$ and $\max \texttt{sep}^- A_i < \max \texttt{sep}^+ A_i = v_i$; since $\preceq$ extends $\leq$, we get $\texttt{sep}^- A_i \prec \texttt{sep}^+ A_i$. Therefore,

$$\{u_1\} = \texttt{sep}^- A_1 \prec \texttt{sep}^+ A_1 = \texttt{sep}^- A_2 \prec \ldots \tag{26}$$
$$\ldots \prec \texttt{sep}^+ A_{k-1} = \texttt{sep}^- A_k \prec \texttt{sep}^+ A_k = \{v_k\}$$

We also have

$$\{u_1\} = \min \mathcal{F}_T \cap \mathcal{S} \qquad \{v_k\} = \max \mathcal{F}_T \cap \mathcal{S} \tag{27}$$

where $\min, \max$ are taken w.r.t. $\preceq$. Indeed, for each $B \in \mathcal{F}_T \cap \mathcal{S}$ there exists $A_i \in \mathcal{O}_T$ with $B \in \mathcal{F}_{A_i}$ (by Assumption 1); using Assumption 6 and the fact that $\preceq$ extends $\leq$, we get $\{u_1\} \prec \{u_2\} \prec \ldots \preceq \{u_i\} \preceq B$. The second equation in (27) is proved in a similar way.

Consider $B \in \mathcal{F}_T \cap \mathcal{S}$. Equations (26),(26) imply that there exists at least one factor $A_i \in \mathcal{O}_T$ with $\texttt{sep}^- A_i \preceq B \preceq \texttt{sep}^+ A_i$. It remains to show that $B \subseteq A_i$; then we will have $B \in \mathcal{F}_{A_i}$ by proposition 2.2, implying $B \in \mathcal{S}_{A_i}$.

Consider node $v \in B$. There holds $\{u_i\} \preceq \texttt{sep}^- A \preceq B$, and therefore $u_i \leq v$. Similarly, $v \leq v_i$. Monotonicity assumption 6 then implies that $v \in A$. The claim is proved.

## Appendix D: proof of theorem 4.2

Averaging $B$ does not affect parameters in trees $T \in \mathcal{T}_B$, so for the purpose of the proof we can assume w.l.o.g. that $\mathcal{T} = \mathcal{T}_B$. Furthermore, we can assume that $\min_{\boldsymbol{x}} \nu^T(\boldsymbol{x}) = 0$ for each $T \in \mathcal{T}_B$ (this can be achieved by adding a constant to $\nu^T(\boldsymbol{x})$; clearly, this does not affect theorem's claims.) We thus have

$$\Phi(\boldsymbol{\theta}) = \sum_{T \in \mathcal{T}_B} \rho^T \min_{\boldsymbol{x}} \nu^T(\boldsymbol{x}) = 0$$

By construction, before the averaging $\nu_B^T$ gives correct min-marginals for $T$ in $B$ (eq. 11). It is easy to see that the same holds after the averaging, i.e. $\tilde\nu_B^T$ gives correct min-marginals for function $f(\cdot \mid \tilde\theta^T)$. (This is because in the definition of min-marginals we fix labeling $\boldsymbol{x}_B$, and vectors $\theta^T, \tilde\theta^T$ differ only in components $\theta_B^T(\boldsymbol{x}_B)$.)

We thus have $\min_{\boldsymbol{x}_B} \nu_B^T(\boldsymbol{x}_B) = \min_{\boldsymbol{x}} \nu^T(\boldsymbol{x}) = 0$. Inspecting update (12), we conclude that $\tilde\nu_B^T(\boldsymbol{x}_B) \geq 0$ for each $\boldsymbol{x}_B$. This gives part (a):

$$\Phi(\tilde{\boldsymbol{\theta}}) = \sum_{T \in \mathcal{T}_B} \rho^T \min_{\boldsymbol{x}} \tilde\nu^T(\boldsymbol{x}) = \sum_{T \in \mathcal{T}_B} \rho_B^T \min_{\boldsymbol{x}_B} \tilde\nu_B^T(\boldsymbol{x}_B) \geq 0$$

To prove part (b), suppose that $\boldsymbol{\theta}$ satisfies RWTA for $B$ with relations $(\mathcal{R}^T \mid T \in \mathcal{T})$. We need to show that $\mathcal{R}^T \subseteq \langle \tilde\nu^T \rangle$ for each $T \in \mathcal{T}_B$. Consider labeling $\boldsymbol{x} \in \mathcal{R}^T$. For each $T' \in \mathcal{T}_B$ we have $\boldsymbol{x}_B \in \pi_B(\mathcal{R}^T) = \pi_B(\mathcal{R}^{T'})$, therefore $\exists \boldsymbol{x}^{T'} \in \langle \nu^{T'} \rangle$ with $\boldsymbol{x}_B^{T'} = \boldsymbol{x}_B$. Since $\nu_B^{T'}$ gives correct min-marginals for $B$ in tree $T'$, we conclude that $\nu_B^{T'}(\boldsymbol{x}_B) = 0$. This implies that $\tilde\nu_B^T(\boldsymbol{x}_B) = 0$ (see eq. 12). This implies that $\boldsymbol{x}_B \in \langle \tilde\nu_B^T \rangle$ and thus $\boldsymbol{x} \in \langle \tilde\nu^T \rangle$.

It remains to prove parts (c,d). We assume from now on that the bound does not change: $\Phi(\tilde{\boldsymbol{\theta}}) = \Phi(\boldsymbol{\theta})$; thus, $\min_{\boldsymbol{x}_B} \tilde\nu_B^T(\boldsymbol{x}_B) = 0$ for $T \in \mathcal{T}$.

Let as fix tree $T \in \mathcal{T}_B$, and let $\boldsymbol{x}$ be a labeling in $\langle \tilde\nu^T \rangle$, so $\boldsymbol{x}_B \in \langle \tilde\nu_B^T \rangle$. We have $\tilde\nu_B^T(\boldsymbol{x}_B) = 0$; inspecting update (12), we conclude that $\nu_B^{T'}(\boldsymbol{x}_B) = 0$ for all $T' \in \mathcal{T}_B$, and so $\boldsymbol{x}_B \in \langle \nu_B^T \rangle$ and $\boldsymbol{x} \in \langle \nu^T \rangle$. This proves that $\langle \tilde\nu^T \rangle \subseteq \langle \nu^T \rangle$.

Now assume that WTA for $B$ does not hold. This means that there exist trees $T, T' \in \mathcal{T}_B$ and labeling $\boldsymbol{x} \in \langle \nu^T \rangle$ such that $\boldsymbol{x}_B \notin \langle \nu_B^{T'} \rangle$. The latter condition means that $\nu_B^{T'}(\boldsymbol{x}_B) > 0$, and therefore $\tilde\nu_B^T(\boldsymbol{x}_B) > 0$, $\boldsymbol{x}_B \notin \langle \tilde\nu_B^T \rangle$ and $\boldsymbol{x} \notin \langle \tilde\nu^T \rangle$. Thus, $\langle \tilde\nu^T \rangle$ is a strict subset of $\langle \nu^T \rangle$.

## Appendix E: proof of theorem 4.5

**Constructing mapping** $\phi : \Omega^* \to \Lambda^*$  The construction will be based on the following lemma.

**Lemma 6.1.** *Consider tree $T \in \mathcal{T}$ and non-empty relation $\mathcal{R} \subseteq \langle \theta^T \rangle$. Vector $\theta^T$ can be reparameterized in such a way that it satisfies*
*(a) $\theta_B^T(\boldsymbol{x}_B) = 0$ for each $B \in \mathcal{F}_T \cap \mathcal{S}$ and each $\boldsymbol{x}_B$;*
*(b) $\pi_A(\mathcal{R}) \subseteq \langle \theta_A^T \rangle$ for each $A \in \mathcal{O}_T$;*
*(c) $\min_{\boldsymbol{x}} f(\boldsymbol{x} \mid \theta^T) = \sum_{A \in \mathcal{O}_T} \min_{\boldsymbol{x}_A} \theta_A^T(\boldsymbol{x}_A)$.*

*Proof.* We use induction on the size of the tree. If $\mathcal{O}_T = \{A\}$ then the claim is straightforward - for each $B \in \mathcal{F}_A$ we just need to "move" parameter $\theta_B^T$ to the outer factor $A \in \mathcal{O}_T$, i.e. update $\theta_A^T(\boldsymbol{x}_A) \mathrel{+}= \theta_B^T(\boldsymbol{x}_B), \theta_B^T(\boldsymbol{x}_B) := 0$.

Now consider the induction step; suppose that $|\mathcal{O}_T| \geq 2$. Let us pick a leaf factor $A \in \mathcal{O}_T$ and do the following. First, reparameterize $\theta^T$ so that $\nu_A^T$ gives correct min-marginals for $A$ in $T$ (eq. 11). Second, for each $B \in \mathcal{F}_A$ "move" all parameters $\theta_B^T$ to $A$, as above. Now consider tree $T'$ obtained from $T$ by removing factor $A$. Vector $\theta^{T'}$ is obtained from $\theta^T$ by setting $\theta_A^{T'} := 0$. Using the fact that $\nu_A^T$ gives correct min-marginals for $A$, we conclude that $\mathcal{R}^{T'} \subseteq \langle \theta^{T'} \rangle$ where we defined $\mathcal{R}^{T'} = \mathcal{R}^T$. Let us now reparameterize $\theta^{T'}$ (together with $\theta^T$) using the induction hypothesis.

By construction, the obtained reparameterization $\theta$ satisfies (a). Since $\theta_A^T = \nu_A^T$ gives correct min-marginals for factor $A$, we have $\pi_A(\langle \theta^T \rangle) = \langle \theta_A^T \rangle$, so property (b) holds for factor $A$. For other factors in $\mathcal{O}_T - \{A\}$ property (b) holds by the induction hypothesis.

To prove (c), we first observe that

$$\sum_{A' \in \mathcal{O}_{T'}} \min_{\boldsymbol{x}_{A'}} \theta_{A'}^{T'}(\boldsymbol{x}_{A'}) \stackrel{(1)}{=} \min_{\boldsymbol{x}} f(\boldsymbol{x} \mid \theta^{T'}) = f(\boldsymbol{x}^* \mid \theta^{T'})$$
$$= f(\boldsymbol{x}^* \mid \theta^T) - \theta_A^T(\boldsymbol{x}^*) \stackrel{(2)}{=} \nu_A^T(\boldsymbol{x}^*) - \theta_A^T(\boldsymbol{x}^*) = 0$$

where $\boldsymbol{x}^*$ is a labeling in $\mathcal{R}^T = \mathcal{R}^{T'}$; (1) holds by the induction hypothesis and (2) holds since by construction $\nu_A^T$ gives correct min-marginals in $T$. We can now write

$$\min_{\boldsymbol{x}} f(\boldsymbol{x} \mid \theta^T) = \min_{\boldsymbol{x}_A} \nu_A^T(\boldsymbol{x}_A) = \min_{\boldsymbol{x}_A} \theta_A^T(\boldsymbol{x}_A) = \sum_{A' \in \mathcal{O}_T} \min_{\boldsymbol{x}_{A'}} \theta_{A'}^T(\boldsymbol{x}_{A'})$$

□

We can now construct mapping $\phi : \Omega^* \to \Lambda^*$. Consider vector $\boldsymbol{\theta} \in \Omega^*$ that satisfies WTA with relations $(\mathcal{R}^T \mid T \in \mathcal{T})$. Let us reparameterize each vector $\theta^T$ as described in lemma 6.1. Clearly, this operation does not affect $\Phi(\boldsymbol{\theta})$, and $\boldsymbol{\theta}$ still satisfies WTA with relations $(\mathcal{R}^T \mid T \in \mathcal{T})$. The result of mapping $\phi$ is now defined as $\theta_A = \sum_{T \in \mathcal{T}} \rho^T \theta^T$. For each $B \in \mathcal{F}$ define $\mathcal{R}_B = \pi_B(\mathcal{R}^T)$ where $T \in \mathcal{T}_B$. (Note, $\mathcal{R}_B$ does not depend on which $T$ is chosen, since WTA holds - see definition 4.1.) It is easy to see that $\theta$ satisfies relaxed $J$-consistency condition with relations $(\mathcal{R}_B \mid B \in \mathcal{F})$. We also have

$$\Phi(\boldsymbol{\theta}) = \sum_{T \in \mathcal{T}} \rho^T \min_{\boldsymbol{x}} f(\boldsymbol{x} \mid \theta^T) = \sum_{T \in \mathcal{T}} \sum_{A \in \mathcal{O}_T} \rho^T \min_{\boldsymbol{x}_A} \theta_A^T(\boldsymbol{x}_A)$$
$$= \sum_{A \in \mathcal{O}} \min_{\boldsymbol{x}_A} \theta_A(\boldsymbol{x}_A) = \sum_{A \in \mathcal{F}} \min_{\boldsymbol{x}_A} \theta_A(\boldsymbol{x}_A) = \Psi(\theta)$$

**Constructing mapping $\psi : \Lambda^* \to \Omega^*$** Consider vector $\theta \equiv \bar{\theta}$ that satisfies the $J$-consistency condition with relations $(\mathcal{R}_B \mid B \in \mathcal{F})$. The argument used in the proof of proposition 2.1 implies that the $\bar{J}$-consistency also holds.

First, let us do the following: for each $B \in \mathcal{S}$ pick outer factor $A \in \mathcal{O}$ with $B \in \mathcal{F}_A$ and "move" vector $\theta_B$ to $A$, i.e. update $\theta_A(\boldsymbol{x}_A) \mathrel{+}= \theta_B(\boldsymbol{x}_B), \theta_B(\boldsymbol{x}_B) := 0$.

**Lemma 6.2.** *The update above does not affect $\Psi(\theta)$, and $\theta$ still satisfies the relaxed $J$-consistency condition with relations $(\mathcal{R}_B \mid B \in \mathcal{F})$.*

*Proof.* Let $\theta$ and $\tilde{\theta}$ be the vectors before and after the update for factors $A \in \mathcal{O}, B \in \mathcal{F}_A - \{A\}$, respectively. Consider labeling $\boldsymbol{x}_A \in \mathcal{R}_A \subseteq \langle \theta_A \rangle$. Note that $\boldsymbol{x}_B \in \pi_B(\mathcal{R}_A) = \mathcal{R}_B \subseteq \langle \theta_B \rangle$. To prove the second claim, we need to show that $\boldsymbol{x}_A \in \langle \tilde{\theta}_A \rangle$. This holds since for any other labelings $\boldsymbol{x}'_A$

$$\tilde{\theta}_A(\boldsymbol{x}_A) = \theta_A(\boldsymbol{x}_A) + \theta_B(\boldsymbol{x}_B) \leq \theta_A(\boldsymbol{x}'_A) + \theta_B(\boldsymbol{x}'_B) = \tilde{\theta}_A(\boldsymbol{x}'_A)$$

The first part holds since

$$\min_{\boldsymbol{x}'_A} \tilde{\theta}_A(\boldsymbol{x}_A) + \min_{\boldsymbol{x}'_B} \tilde{\theta}_B(\boldsymbol{x}_B) = \tilde{\theta}_A(\boldsymbol{x}_A) + \tilde{\theta}_B(\boldsymbol{x}_B)$$
$$= \theta_A(\boldsymbol{x}_A) + \theta_B(\boldsymbol{x}_B) = \min_{\boldsymbol{x}'_A} \theta_A(\boldsymbol{x}_A) + \min_{\boldsymbol{x}'_B} \theta_B(\boldsymbol{x}_B)$$

$\square$

We now have vector $\theta$ with $\theta_B(\boldsymbol{x}_B) = 0$ for all $B \in \mathcal{S}$.

**Lemma 6.3.** *Consider tree $T = (\mathcal{O}_T, \mathcal{E}_T)$. Define vector $\theta^T$ as follows: $\theta_A^T = \frac{1}{\rho^T} \theta_A$ for $A \in \mathcal{O}_T$ and $\theta_B^T(\boldsymbol{x}_B) = 0$ for $B \in \mathcal{S}, T \in \mathcal{T}_B$. Define relation*

$$\mathcal{R}^T = \{\boldsymbol{x} \mid \boldsymbol{x}_A \in \mathcal{R}_A \quad \forall A \in \mathcal{O}_T\} \tag{28}$$

*(a) $\pi_B(\mathcal{R}^T) = \mathcal{R}_B$ for each $B \in \mathcal{F}_T$*
*(b) $f(\boldsymbol{x}^T \mid \theta^T) = \sum_{A \in \mathcal{O}_T} \min_{\boldsymbol{x}_A} \theta_A^T(\boldsymbol{x}_A)$ for each $\boldsymbol{x}^T \in \mathcal{R}^T$.*
*(c) $\mathcal{R}^T \subseteq \langle \nu^T \rangle$.*

*Proof.* It suffices to show that $\pi_A(\mathcal{R}^T) = \mathcal{R}_A$ for each $A \in \mathcal{O}_T$; for $B \in \mathcal{F}_A - \{A\}$ we will then have $\pi_B(\mathcal{R}^T) = \pi_B(\mathcal{R}_A) = \mathcal{R}_B$, where the last equality holds since $(A, B) \in \bar{J}$ and $\theta$ satisfies the $\bar{J}$-consistency condition with relations $(\mathcal{R}_B \mid B \in \mathcal{F})$.

We use induction on the size of the tree. For $\mathcal{O}_T = \{A\}$ the claim is obvious; suppose that $|\mathcal{O}_T| \geq 2$. Pick a leaf factor $A \in \mathcal{O}_T$, with $(A, \hat{A}) \in \mathcal{E}_T$. Let $T'$ be the tree obtained from $T$ by removing factor $A$, and $S = A \cap \hat{A} \in \mathcal{F}_{T'}$. We assume that $\rho^{T'} = \rho^T$. By the running intersection property, $(A - S) \cap A' = \varnothing$ for $A' \in \mathcal{O}_T - \{A\}$.

Let $\boldsymbol{x}'$ be a labeling in $\mathcal{R}^{T'}$. By the induction hypothesis $\boldsymbol{x}'_S \in \mathcal{R}_S$. Let $\boldsymbol{x}_A$ be labeling in $\mathcal{R}_A$ with $\boldsymbol{x}_S = \boldsymbol{x}'_S$ (it exists since $(A, S) \in \bar{J}$ and $\bar{J}$-consistency holds). Let $\boldsymbol{x}$ be the labeling obtained from $\boldsymbol{x}'$ by changing the labeling of $A - S$ from $\boldsymbol{x}'_{A-S}$ to $\boldsymbol{x}_{A-S}$. Clearly, $\boldsymbol{x} \in \mathcal{R}^T$.

The argument above and the induction hypothesis show that $\mathcal{R}_{A'} \subseteq \pi_{A'}(\mathcal{R}^T)$ for each $A' \in \mathcal{O}_T - \{A\}$. The fact that $\mathcal{R}_A \subseteq \pi_A(\mathcal{R}^T)$ is also clear (in the argument above we can first choose $\boldsymbol{x}_A \in \mathcal{R}_A$, and then $\boldsymbol{x}' \in \mathcal{R}^{T'}$ which is consistent with $\boldsymbol{x}$ on $S$). The inclusion $\pi_{A'}(\mathcal{R}^T) \subseteq \mathcal{R}_{A'}$ for $A' \in \mathcal{O}_T$ follows from the definition of $\mathcal{R}^T$. This proves part (a). Part (b) is also easy to prove: for each $\boldsymbol{x}^T \in \mathcal{R}^T$ we have

$$f(\boldsymbol{x}^T \mid \theta^T) = \sum_{A' \in \mathcal{O}_T} \theta_{A'}^T(\boldsymbol{x}_{A'}^T) \stackrel{(1)}{=} \sum_{A' \in \mathcal{O}_T} \min_{\boldsymbol{x}_{A'}} \theta_{A'}^T(\boldsymbol{x}_{A'})$$

where (1) holds by the induction hypothesis and the fact that $\boldsymbol{x}_A^T \in \mathcal{R}_A$. Finally, part (c) follows from (b) and the fact that $\sum_{A \in \mathcal{O}_T} \min_{\boldsymbol{x}_A} \theta_A^T(\boldsymbol{x}_A)$ is a lower bound on $\min_{\boldsymbol{x}} f(\boldsymbol{x} \mid \theta^T)$. $\square$

The result of mapping $\psi$ is now defined as described in the lemma. It is easy to see that the obtained vector $\boldsymbol{\theta}$ satisfies WTA with relations $(\mathcal{R}^T \mid T \in \mathcal{T})$ from the lemma. We also have

$$\Phi(\boldsymbol{\theta}) = \sum_{T \in \mathcal{T}} \rho^T \min_{\boldsymbol{x}^T} \nu^T(\boldsymbol{x}^T) \stackrel{(1)}{=} \sum_{T \in \mathcal{T}} \sum_{A \in \mathcal{O}_T} \rho^T \min_{\boldsymbol{x}_A} \theta_A^T(\boldsymbol{x}_A) = \Psi(\boldsymbol{\theta})$$

where (1) follows from lemma 6.3(b,c).